\documentclass{article} 
\usepackage{iclr2026_conference,times}

\usepackage{amsmath,amsfonts,bm}

\def\eqref#1{equation~\ref{#1}}

\def\1{\bm{1}}

\DeclareMathAlphabet{\mathsfit}{\encodingdefault}{\sfdefault}{m}{sl}
\SetMathAlphabet{\mathsfit}{bold}{\encodingdefault}{\sfdefault}{bx}{n}

\usepackage{hyperref}
\usepackage{url}
\usepackage{graphicx} 

\usepackage{booktabs}
\usepackage{tabularx}
\usepackage{siunitx}

\usepackage{amssymb}

\usepackage{booktabs,tabularx,makecell}

\title{Uncertainty-Gated Generative Modeling}

\author{
Xingrui Gu\thanks{Equal contribution.} \\
Electrical Engineering and Computer Science \\
University of California, Berkeley \\
Berkeley, CA, USA \\
\texttt{xingrui\_gu@berkeley.edu}
\And
Haixi Zhang\footnotemark[1] \\
School of Electrical and Computer Engineering \\
Cornell University \\
Ithaca, NY, USA \\
\texttt{hz733@cornell.edu}
}

\iclrfinalcopy

\begin{document}

\maketitle

\begin{abstract}
Financial time-series forecasting is a high-stakes problem where regime shifts and shocks make point-accurate yet overconfident models dangerous. We propose Uncertainty-Gated Generative Modeling (UGGM), which treats uncertainty as an internal control signal that gates (i) representation via gated reparameterization, (ii) propagation via similarity × confidence routing, and (iii) generation via uncertainty-controlled predictive distributions, together with uncertainty-driven regularization and calibration to curb miscalibration. Instantiated on Weak Innovation AutoEncoder (WIAE-GPF), our UG-WIAE-GPF significantly improves risk-sensitive forecasting, delivering a 63.5\% MSE reduction on NYISO (0.3508 $\rightarrow$ 0.1281), with improved robustness under shock intervals (mSE: 0.2739 $\rightarrow$ 0.1748). 
\end{abstract}

\section{Introduction}
Financial time-series forecasting is more than minimizing error: forecasts directly drive trading, hedging, liquidity/margin management, and risk limits, so small biases can compound under non-stationarity into outsized losses. Regime shifts make distributional drift an expected failure mode \citep{hamilton1989new}, and heavy tails, jumps, and volatility clustering make extremes far more likely than Gaussian assumptions imply \citep{cont2001empirical, embrechts2013modelling}. Meanwhile, governance and regulatory accountability require interpretable, calibrated risk measures, not just point forecasts \citep{kiritz2018supervisory, basel2013principles, best2021minimum}. Hence, models for financial agents should produce uncertainty-aware forecasts and conservative inference under rising uncertainty; otherwise, point-estimation training can yield miscalibration and overconfidence precisely when errors are most costly \citep{gneiting2007probabilistic}.


Traditional statistical baselines like AutoARIMA and AutoETS offer transparency but struggle with nonlinearity, heavy-tailed noise, and structural breaks \citep{hyndman2008automatic, hamilton1989new}. While deep learning models, such as PatchTST and TimesNet, enhance expressiveness \citep{nie2022time, wu2022timesnet}, attention-based architectures typically rely on similarity-routed aggregation \citep{vaswani2017attention, gu2024advancing}. These methods lack principled mechanisms to modulate trust in evidence, often amplifying spurious signals into overconfident forecasts during regime shifts. Furthermore, most models prioritize point accuracy over the calibrated, decision-relevant uncertainty required for high-stakes deployment and regulatory compliance \citep{gneiting2007probabilistic, basel2013principles, gu2025causkelnet, gu2024mimicking, gu2026laplacian}.


To bridge this gap, we propose \emph{Uncertainty-Gated Generative Modeling} (UGGM), which turns uncertainty into an internal control signal that gates representation, propagation, and generation. UGGM follows a three-stage pipeline: (i) \textbf{Gated representation}, with encoder outputs $(\mu,\sigma)$ and gate $g\in[0,1]$ defining $z=\mu+g,\epsilon\odot\sigma$; (ii) \textbf{Gated propagation}, weighting interactions by similarity $\times$ gated confidence to downweight unreliable evidence; and (iii) \textbf{Gated generation}, producing uncertainty-conditioned predictive distributions. With uncertainty-aware regularization and calibration losses, UG-WIAE (UGGM on WIAE) yields $63.5\%$ MSE reduction on NYISO price forecasting and improves robustness under regime shifts and shocks.

\section{Preliminaries}

Let $\mathbf{x}_{1:T}\in\mathbb{R}^{T\times D}$ be a multivariate time series with $\mathbf{x}_t\in\mathbb{R}^D$. 
At each time $t$, we use a rolling window with history length $L$ and horizon $H$:
\begin{align}
\mathbf{X}_t := \mathbf{x}_{t-L+1:t}\in\mathbb{R}^{L\times D},\qquad
\mathbf{Y}_t := \mathbf{x}_{t+1:t+H}\in\mathbb{R}^{H\times D},
\end{align}
where $\mathbf{X}_t$ is the context and $\mathbf{Y}_t$ the future target. 
We learn a conditional generative forecaster $p_\theta(\mathbf{Y}_t\mid \mathbf{X}_t)$ that models stochastic future trajectories and yields calibrated predictive uncertainty for reliable decision-making \citep{gneiting2007probabilistic}.


\section{Uncertainty-Gated Generative Modeling}
Financial time series exhibit non-stationarity, heavy-tailed shocks, and structural breaks, where point forecasts can induce decision-level overconfidence by remaining highly confident precisely during regime shifts or anomalies. In high-risk settings, forecasting should therefore ``guess intelligently'': align errors with uncertainty and use uncertainty to regulate model behavior for improved risk controllability.

\paragraph{Uncertainty decomposition.}
We summarize predictive uncertainty into three interpretable components:
\begin{equation}
U_t^{\mathrm{total}} \;=\; U_t^{\mathrm{data}} \;\oplus\; U_t^{\mathrm{model}} \;\oplus\; U_t^{\mathrm{dec}},
\end{equation}
where $\oplus$ denotes a conceptual aggregation. Data uncertainty captures irreducible noise,
\begin{equation}
U_t^{\mathrm{data}} \triangleq \mathbb{E}\!\left[\mathrm{Var}\!\left(\mathbf{Y}_t \mid \mathbf{X}_t\right)\right],
\end{equation}
model uncertainty captures epistemic disagreement under finite data,
\begin{equation}
U_t^{\mathrm{model}} \triangleq \mathrm{Var}\!\left(\mathbb{E}\!\left[f_\theta(\mathbf{X}_t)\right]\right),
\end{equation}
and decision uncertainty parameterizes conservatism via a learnable gate $g_t\in[0,1]$,
\begin{equation}
U_t^{\mathrm{dec}} \triangleq \mathrm{Var}(g_t).
\end{equation}

\subsection{Uncertainty-Gated Principle (UG): uncertainty as control}
Most forecasters surface uncertainty only at the output (e.g., intervals or distribution parameters). UG instead elevates uncertainty to a \emph{control signal} that modulates internal computation.
Let $\mathbf{c}_t$ denote optional context. We define a scalar gate
\begin{align}
\mathbf{u}_t &:= \psi_\eta\!\Big(\big[U_t^{\mathrm{data}},\,U_t^{\mathrm{model}},\,\mathbf{c}_t\big]\Big), \\
g_t &:= \sigma\!\left(W_g\,\mathbf{u}_t + b_g\right)\in[0,1],
\label{eq:gate}
\end{align}
where $[\cdot]$ is concatenation and $\psi_\eta$ is a small MLP (or identity). Intuitively, smaller $g_t$ corresponds to more conservative behavior under higher uncertainty.

\paragraph{UG actions.}
UG specifies a family of \emph{gated actions} that turn uncertainty into risk-sensitive computation. In this work we instantiate three actions:
\begin{gather}
\textbf{(A) Gated stochasticity:}\quad
\mathbf{z}_t
= \boldsymbol{\mu}_t + g_t\,\boldsymbol{\epsilon}_t\odot\boldsymbol{\sigma}_t,
\qquad \boldsymbol{\epsilon}_t\sim\mathcal{N}(\mathbf{0},\mathbf{I}),
\label{eq:ug_action_latent}
\\[0.6em]
\textbf{(B) Relevance$\times$confidence routing:}\quad
A_{ij}
\propto \exp\!\Big(\tfrac{\mathbf{q}_i^\top\mathbf{k}_j}{\sqrt{d}}\Big)\cdot
G(\sigma_i^{(q)},\sigma_j^{(k)}),
\label{eq:ug_action_route}
\\[0.6em]
\textbf{(C) Adaptive conservatism/regularization:}\quad
\lambda_t
= \lambda_0\,\phi(g_t)\quad\text{(e.g., $\phi(g)=1-g$)},
\label{eq:ug_action_reg}
\end{gather}

where (A) injects uncertainty-dependent stochasticity into latent representations, (B) down-weights low-confidence evidence in attention-based aggregation, and (C) adjusts model conservatism (e.g., regularization strength or inference smoothing) as uncertainty increases.
This makes uncertainty actionable across \emph{representation, aggregation, and generation}.

\subsection{Instantiation: probabilistic encoding, UG transformation, probabilistic generation}
We implement uncertainty-gated (UG) actions on top of any forecasting backbone with three modules: (i) a probabilistic encoder with gated reparameterization, (ii) a UG-modulated transformation (including UG attention when used), and (iii) a probabilistic decoder producing a predictive distribution.

\paragraph{Probabilistic encoder.}
Given $\mathbf{X}_t\in\mathbb{R}^{L\times D}$, the encoder outputs
\begin{equation}
\mathbf{h}_t=\mathrm{Enc}_\theta(\mathbf{X}_t),\qquad
\boldsymbol{\mu}_t=W_\mu\mathbf{h}_t+\mathbf{b}_\mu,\qquad
\log\boldsymbol{\sigma}_t^2=W_\sigma\mathbf{h}_t+\mathbf{b}_\sigma,
\end{equation}
and samples the latent state via the UG gate $g_t$ (Eq.~\eqref{eq:gate}):
\begin{equation}
\mathbf{z}_t=\boldsymbol{\mu}_t+g_t\,\boldsymbol{\sigma}_t\odot\boldsymbol{\epsilon}_t,\qquad
\boldsymbol{\epsilon}_t\sim\mathcal{N}(\mathbf{0},\mathbf{I}).
\label{eq:gated_latent}
\end{equation}

\paragraph{Uncertainty-gated attention (relevance$\times$confidence routing; Eq.~\eqref{eq:ug_action_route}).}
For attention backbones, we gate interactions by confidence. With
$\mathbf{q}_i=W_q\mathbf{z}_i$, $\mathbf{k}_j=W_k\mathbf{z}_j$, $\mathbf{v}_j=W_v\mathbf{z}_j$,
\begin{equation}
A_{ij}=\mathrm{Softmax}_j\!\left(\frac{\mathbf{q}_i^\top\mathbf{k}_j}{\sqrt{d}}+\log G(\sigma_i^{(q)},\sigma_j^{(k)})\right),
\qquad
\tilde{\mathbf{z}}_i=\sum_j A_{ij}\mathbf{v}_j,
\label{eq:ug_attn}
\end{equation}
where we use
\begin{equation}
G(\sigma_i^{(q)},\sigma_j^{(k)})=\exp\!\left(-\alpha(\sigma_i^{(q)}+\sigma_j^{(k)})\right),\quad \alpha\ge 0,
\label{eq:conf_gate}
\end{equation}
which down-weights low-confidence pairs.

\paragraph{Probabilistic decoder (predictive distribution and sharpness control).}
The decoder outputs distribution parameters
\begin{equation}
(\boldsymbol{\mu}_t^{y},\boldsymbol{\sigma}_t^{y})=\mathrm{Dec}_\theta(\tilde{\mathbf{z}}_t),\qquad
\mathbf{Y}_t\mid\mathbf{X}_t\sim\mathcal{N}\!\left(\boldsymbol{\mu}_t^{y},\,\mathrm{diag}\!\big((\boldsymbol{\sigma}_t^{y})^2\big)\right),
\label{eq:pred_dist}
\end{equation}
and, when sampling,
\begin{equation}
\hat{\mathbf{Y}}_t=\boldsymbol{\mu}_t^{y}+g_t'\,\boldsymbol{\sigma}_t^{y}\odot\boldsymbol{\epsilon}_t',\qquad
\boldsymbol{\epsilon}_t'\sim\mathcal{N}(\mathbf{0},\mathbf{I}),
\label{eq:pred_sample}
\end{equation}
where $g_t'\in[0,1]$ controls output sharpness (default $g_t'=g_t$, or predict $g_t'$ separately).

\subsection{Training objective}
We optimize
\begin{equation}
\mathcal{L}=\mathcal{L}_{\mathrm{NLL}}+\lambda_1\mathcal{L}_{\mathrm{cal}}+\lambda_2\mathcal{L}_{\mathrm{gate}}.
\label{eq:loss_total}
\end{equation}
Gaussian NLL:
\begin{equation}
\mathcal{L}_{\mathrm{NLL}}=
\frac{1}{N}\sum_{n=1}^{N}\sum_{h=1}^{H}
\left[
\frac{(y_{n,h}-\mu_{n,h}^{y})^2}{2(\sigma_{n,h}^{y})^2}
+\frac{1}{2}\log (\sigma_{n,h}^{y})^2
\right].
\label{eq:nll}
\end{equation}
Calibration alignment:
\begin{equation}
\mathcal{L}_{\mathrm{cal}}=\left(1-\mathrm{Corr}\!\left(\left|\mathbf{Y}-\boldsymbol{\mu}^{y}\right|,\boldsymbol{\sigma}^{y}\right)\right)^2.
\label{eq:cal}
\end{equation}
Gate smoothness:
\begin{equation}
\mathcal{L}_{\mathrm{gate}}=\sum_t\|g_t-g_{t-1}\|_2^2.
\label{eq:gate_smooth}
\end{equation}

\subsection{Risk-sensitive inference}
At inference, the model outputs $p_\theta(\mathbf{Y}_t\mid\mathbf{X}_t)$ with parameters $(\boldsymbol{\mu}_t^{y},\boldsymbol{\sigma}_t^{y})$ (and optionally a gate $g_t$). We summarize uncertainty by a scalar risk score
\begin{equation}
r_t := \rho(\boldsymbol{\sigma}_t^{y})\in\mathbb{R}_{+},
\end{equation}
for a fixed functional $\rho(\cdot)$ (e.g., $\|\boldsymbol{\sigma}\|_2$ or $\max_i\sigma_i$), and select an inference action $a_t\in\mathcal{A}$ by thresholding:
\begin{equation}
a_t =
\begin{cases}
a^{\mathrm{std}}, & r_t \le \tau,\\
a^{\mathrm{rob}}, & r_t > \tau.
\end{cases}
\label{eq:action_rule}
\end{equation}
The conservative action $a^{\mathrm{rob}}$ can be instantiated by uncertainty inflation $\boldsymbol{\sigma}_t^{y}\!\leftarrow\!\kappa\boldsymbol{\sigma}_t^{y}$ ($\kappa>1$), increased sampling budget $S\!\leftarrow\!S_{\mathrm{rob}}$, and/or stronger smoothing $\boldsymbol{\mu}_t^{y}\!\leftarrow\!\mathcal{S}_\beta(\boldsymbol{\mu}_t^{y})$, yielding an explicit mapping from predictive uncertainty to risk-aware inference. We implement the proposed UG framework on the WIAE-GPF model\citep{wang2024probabilistic}; implementation details are provided in the Appendix~\ref{sec:appendix_wiae}.

\section{Experiment}

\begin{table}[t]
\centering
\caption{Comparison with state-of-the-art forecasters on NYISO (2018--2024). Lower is better. \textbf{Best} is bold; \underline{second best} is underlined (ties share the same rank).}
\label{tab:results}
\vspace{0.4em}
\setlength{\tabcolsep}{4.2pt}
\renewcommand{\arraystretch}{1.12}
\begin{small}
\begin{tabularx}{\linewidth}{l*{8}{>{\centering\arraybackslash}X}}
\toprule
& \multicolumn{2}{c}{\textbf{Pointwise}} 
& \multicolumn{2}{c}{\textbf{Normalized}} 
& \multicolumn{2}{c}{\textbf{Robust (Med.)}} 
& \multicolumn{2}{c}{\textbf{Scale-free}} \\
\cmidrule(lr){2-3}\cmidrule(lr){4-5}\cmidrule(lr){6-7}\cmidrule(lr){8-9}
Model &
\multicolumn{1}{c}{MSE $\downarrow$} &
\multicolumn{1}{c}{MAE $\downarrow$} &
\multicolumn{1}{c}{NMSE $\downarrow$} &
\multicolumn{1}{c}{NMAE $\downarrow$} &
\multicolumn{1}{c}{mSE $\downarrow$} &
\multicolumn{1}{c}{mAE $\downarrow$} &
\multicolumn{1}{c}{MAPE $\downarrow$} &
\multicolumn{1}{c}{MASE $\downarrow$} \\

\midrule

\multicolumn{9}{l}{\textit{Proposed method (WIAE-based)}} \\
\textbf{UG-WIAE-GPF (Ours)} &
\textbf{0.1281} & \textbf{0.2550} &
\underline{0.0119} & \underline{0.0280} &
\textbf{0.1748} & \textbf{0.2145} &
\textbf{0.2515} & \textbf{0.0220} \\
WIAE-GPF (2024) &
\underline{0.3508} & \underline{0.3835} &
0.0220 & 0.0399 &
\underline{0.2739} & \underline{0.3466} &
\underline{0.3958} & 0.3990 \\
\midrule

\multicolumn{9}{l}{\textit{Transformer-/decomposition-based baselines}} \\
PatchTST (2022)   & 1.7508 & 1.0866 & 0.0387 & 0.0320 & 10.5493 & 3.2480 & 3.3637 & 0.1648 \\
NBEATS (2019)     & 2.3578 & 0.9501 & 0.0521 & \underline{0.0280} & 11.1128 & 3.3216 & 2.4569 & 0.1441 \\
NHITS (2023)      & 2.7831 & 1.0988 & 0.0615 & 0.0324 & 12.0617 & 3.4727 & 2.9154 & 0.1666 \\
TimesNet (2023)   & 6.3864 & 1.9325 & 0.1430 & 0.0571 & 11.8695 & 3.4448 & 5.4370 & 0.2930 \\
TimeMixer (2024)  & 9.1532 & 1.8871 & 0.2024 & 0.0556 & 15.1240 & 3.8874 & 4.9194 & 0.2862 \\
TFT (2021)        & 23.4326 & 3.4796 & 0.5181 & 0.1025 & 13.1885 & 3.6304 & 9.1168 & 0.5277 \\
DeepAR (2020)     & 452.31 & 20.176 & 10.000 & 0.5941 & 374.87 & 19.362 & 57.887 & 3.0595 \\
\midrule

\multicolumn{9}{l}{\textit{Recurrent neural network baselines}} \\
GRU (2017)        & 0.4124 & 0.4245 & \textbf{0.0092} & \textbf{0.0126} & 10.4911 & 3.2388 & 1.1081 & \underline{0.0644} \\
RNN (2014)        & 3.6939 & 1.6508 & 0.0827 & 0.0488 & 15.6077 & 3.9506 & 4.6400 & 0.2503 \\
LSTM (2014)       & 4.2307 & 1.8180 & 0.0947 & 0.0538 & 18.0323 & 4.2464 & 5.1167 & 0.2757 \\
DilatedRNN (2017) & 6.1914 & 1.7639 & 0.1386 & 0.0522 & 8.3932 & 2.8971 & 4.8401 & 0.2675 \\
\midrule

\multicolumn{9}{l}{\textit{Lightweight / linear / graph baselines}} \\
MLP (2010)        & 14.3665 & 2.6669 & 0.3176 & 0.0785 & 5.9389 & 2.4364 & 7.1342 & 0.4044 \\
NLinear (2023)    & 45.3496 & 4.9876 & 1.0156 & 0.1475 & 21.5241 & 4.6061 & 13.9908 & 0.7563 \\
StemGNN (2020)    & 54.2191 & 5.6601 & 1.2142 & 0.1674 & 19.6572 & 4.4332 & 15.8152 & 0.8583 \\
\bottomrule
\end{tabularx}
\end{small}
\vspace{-0.6em}
\end{table}

We conduct 24-hour (1 day) ahead probabilistic forecasting on the NYISO univariate load series, using a rolling window of the past 168 hours (7 days) as input, all benchmark model from NeuralForecast library\citep{olivares2022library_neuralforecast}. The data are split chronologically, with out-of-sample evaluation starting from 2023-01-01. For each input, we draw 500 samples to estimate the predictive distribution; the mean and median are reported as point forecasts and evaluated using MSE, MAE, and their normalized variants. As shown in Table~\ref{tab:results}, UG-WIAE-GPF achieves the best overall performance on the NYISO (2018--2024) in \ref{dataset} 24-hour forecasting task, ranking first across point errors (MSE/MAE), robust statistics (mSE/mAE), and scale-free metrics (MAPE/MASE), and substantially reducing error relative to the ungated uncertainty baseline (WIAE). Notably, UG-WIAE exhibits larger gains in median-based errors, indicating greater stability under anomalous fluctuations and heavy-tailed segments. At the same time, it remains leading or near-optimal on normalized errors (NMSE/NMAE), suggesting that the improvements are not driven by scale effects but by more reliable distributional generation and risk-sensitive forecasting. In the end, UG-WIAE-GPF treats uncertainty as a control signal that gates representation, attention, and generation, enabling risk-sensitive forecasts and delivering strong NYISO gains (e.g., 63.5\% MSE reduction).

\section{Conclusion}
We propose Uncertainty-Gated (UG) probabilistic forecasting, treating uncertainty as a control signal that modulates representation, information routing, and generation. On NYISO (2018–2024) 24-hour forecasting, UG-WIAE-GPF markedly improves over WIAE-GPF, reducing MSE 0.3508→0.1281 and MAE 0.3835→0.2550, while strengthening robustness (mSE 0.1748, mAE 0.2145) and achieving the best scale-free accuracy (MAPE 0.2515, MASE 0.0220) with competitive normalized errors (NMSE 0.0119, NMAE 0.0280). Overall, uncertainty-as-control yields more stable, interpretable distributional forecasts in high-stakes, non-stationary settings, supporting risk-aware agentic decision-making.

\newpage
\bibliography{iclr2026_conference}
\bibliographystyle{iclr2026_conference}
\newpage
\appendix
\section{Appendix}

\subsection{Dataset Detail}
\label{dataset}
\begin{figure}[h]
    \centering
    \includegraphics[width=\linewidth]{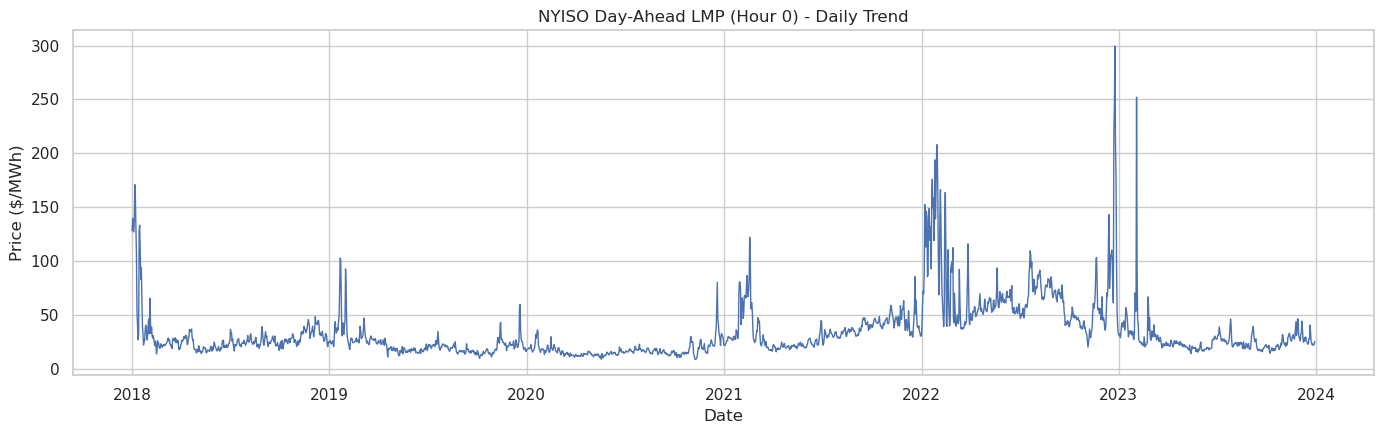}
    \vspace{0.6em}
    \includegraphics[width=\linewidth]{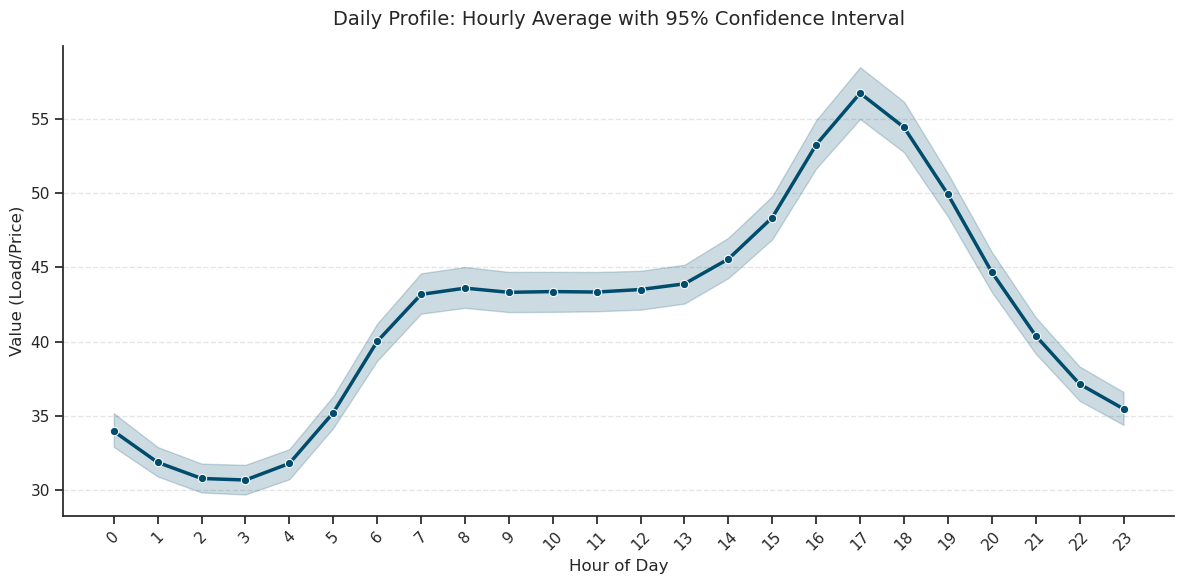}
    \caption{
    \textbf{NYISO data characteristics.}
    \emph{Top:} Long-term day-ahead LMP trend (2018--2024) showing strong non-stationarity and extreme spikes.
    \emph{Bottom:} Daily profile with hourly averages and 95\% confidence intervals, highlighting pronounced intra-day heteroskedasticity.
    }
    \label{fig:nyiso_appendix}
\end{figure}

\begin{figure}[h]
  \centering
  \includegraphics[width=0.85\linewidth]{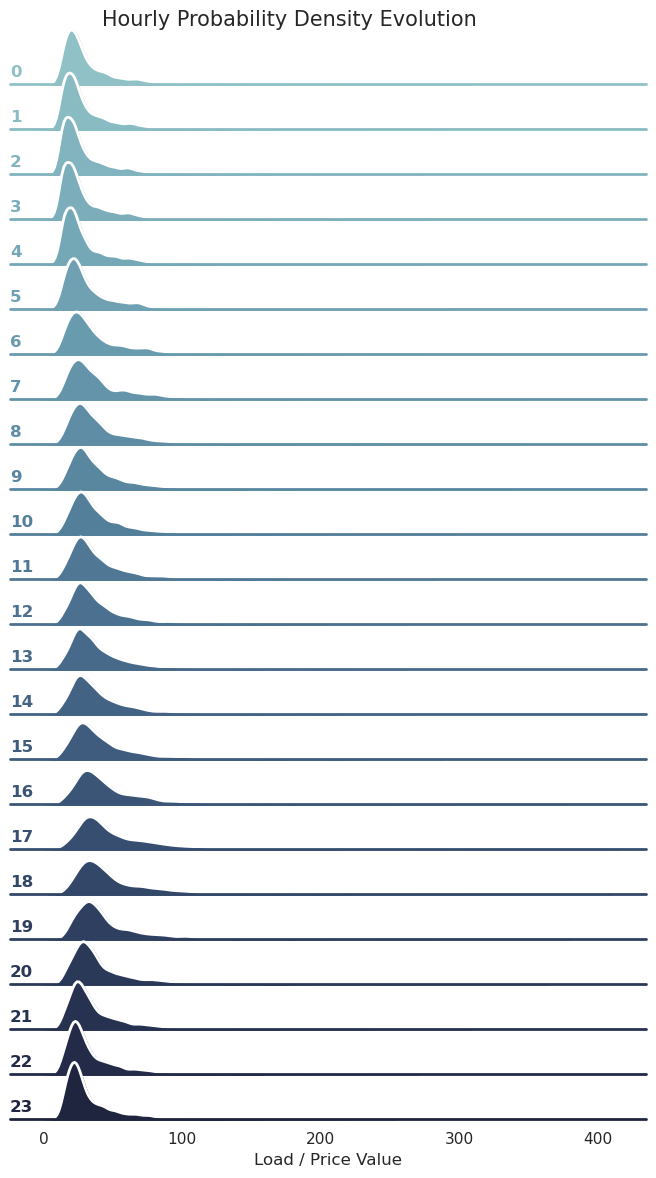}
  \caption{Hourly probability density evolution of NYISO load/price. 
  Distributions vary substantially across hours in location, scale, and tail mass, revealing strong intra-day heteroskedasticity and heavy-tailed extremes.}
  \label{fig:nyiso_hourly_density}
  \vspace{-0.6em}
\end{figure}

NYISO exhibits the key properties that motivate UGGM: pronounced non-stationarity
(regime changes over 2018--2024), frequent heavy-tailed spikes and outliers
(extreme price/load events), and strong intra-day heteroskedasticity in Fig.~\ref{fig:nyiso_appendix}.
As shown in Fig.~\ref{fig:nyiso_hourly_density}, the hourly probability density
evolves markedly across the day, with substantial shifts in location, dispersion,
and tail behavior. This hour-dependent distributional drift concentrates error
and risk in rare but high-impact periods, making point forecasting brittle and
exposing miscalibration. Consequently, effective models must go beyond mean
prediction and adapt their behavior under uncertainty, producing calibrated
distributional forecasts and robust aggregation. NYISO therefore serves as a
stress-test domain where \emph{uncertainty-as-control} is directly relevant for
risk-sensitive forecasting.

\newpage
\newpage
\subsection{Weak Innovation Autoencoder (WIAE): Background and Formulation}
\label{sec:appendix_wiae}

\paragraph{Terminology.} We use \emph{Weak Innovation Autoencoder} (WIAE) to denote an autoencoding construction that enables \emph{weak innovation representations} for nonparametric time series\citep{wang2024probabilistic}. We refer to the resulting probabilistic forecasting pipeline as \emph{WIAE-based Generative Probabilistic Forecasting} (WIAE-GPF).

\subsubsection{Notation}
Let $(X_t)_{t\ge 0}$ be the target time-series process and $(V_t)_{t\ge 0}$ an \emph{innovation} (exogenous noise) process. Let $(U_t)_{t\ge 0}$ be an i.i.d.\ reference process with uniform marginals (typically $U_t\sim \mathrm{Unif}([0,1]^m)$ for some dimension $m$). The WIAE consists of an encoder $G$ and decoder $H$, parameterized as $G_\theta$ and $H_\eta$. We denote by $\hat V_t$ the encoded innovation extracted from data, and by $\hat X$ the reconstructed/generated sequence produced by the decoder.

\subsubsection{Weak innovation representation and WIAE}
Classical \emph{strong} innovation representations (with strict pointwise invertibility/innovation recovery) need not exist for general nonparametric time series. Rosenblatt therefore introduced a \emph{weak innovation representation}, which only requires \emph{distributional} consistency between (i) the observed past concatenated with a generated future and (ii) the true joint trajectory.

Fix a forecasting horizon $T$. A pair $(G,H)$ is said to realize a weak innovation representation if, for any time index $t$,
\begin{equation}
\big(X_{0:t},\,\hat X_{t+1:t+T}\big)\;\overset{d}{=}\;X_{0:t+T},\qquad \forall\,t,
\label{eq:weak_innovation_rep}
\end{equation}
where $\hat X_{t+1:t+T}$ is the decoder output conditioned on the observed past $X_{0:t}$. The original work calls any autoencoder $(G,H)$ satisfying~\eqref{eq:weak_innovation_rep} a \emph{Weak Innovation Autoencoder} (WIAE).

\paragraph{Intuition.}
Unlike standard autoencoders that minimize pointwise reconstruction error, WIAE enforces \emph{joint distribution matching}: the concatenated sequence ``past + generated future'' should be indistinguishable (in distribution) from the true ``past + true future''. This provides a principled foundation for \emph{generative} probabilistic forecasting, where the objective is to recover the correct conditional law of future trajectories given the past.

\subsubsection{Learning WIAE via dual discriminators}
To enforce both (i) the desired innovation structure and (ii) the weak distribution matching in~\eqref{eq:weak_innovation_rep}, WIAE training introduces two discriminators:
\begin{itemize}
\item \textbf{Innovation discriminator} $D_\gamma$, which compares the distribution of extracted innovations $(\hat V_t)$ to the i.i.d.\ uniform reference $(U_t)$;
\item \textbf{Reconstruction discriminator} $D_\omega$, which compares the true joint trajectory $X_{0:t+T}$ to the concatenated pair $(X_{0:t},\hat X_{t+1:t+T})$.
\end{itemize}
The original formulation uses Wasserstein critics and combines the two discrepancies into a single minimax objective:
\begin{equation}
\min_{\theta,\eta}\;\max_{\gamma,\omega}\;
\Big(\mathbb{E}\big[D_\gamma(U_t)\big]-\mathbb{E}\big[D_\gamma(\hat V_t)\big]\Big)
\;+\;
\lambda\Big(\mathbb{E}\big[D_\omega(X_{0:t+T})\big]-\mathbb{E}\big[D_\omega(X_{0:t},\hat X_{t+1:t+T})\big]\Big),
\label{eq:wiae_minimax}
\end{equation}
where $\lambda>0$ balances innovation matching and reconstruction matching.

WIAE-GPF can be viewed as sampling-based forecasting: given an observed history $x_{0:t}$, one samples i.i.d.\ uniform exogenous noise and uses the learned decoder to generate future trajectories. Under ideal learning, the generated future is distributed according to the true conditional law, i.e.,
\begin{equation}
\hat X_{t+1:t+T}\;\Big|\;X_{0:t}=x_{0:t}
\;\sim\;
\mathcal{L}\!\left(X_{t+1:t+T}\,\big|\,X_{0:t}=x_{0:t}\right),
\label{eq:wiae_gpf_semantics}
\end{equation}
which captures the core semantic behind the ``provably correct'' claim in WIAE-GPF: correctness is defined as recovering the \emph{conditional distribution} of future trajectories rather than only minimizing point errors.

\subsection{UG principle: uncertainty as an internal control signal}

The central idea of UG-WIAE-GPF is to elevate uncertainty from a passive output-level quantity to an \emph{operational control signal} that permeates the entire \emph{encoding–aggregation–generation} pipeline. Without altering the adversarial training paradigm of WIAE—where distributional consistency is enforced by innovation and reconstruction discriminators—we introduce three plug-and-play mechanisms inside the generator:  
(i) probabilistic encoding with gated noise injection,  
(ii) uncertainty-aware attention, and  
(iii) probabilistic decoding with gated sampling.  
Together, these mechanisms allow uncertainty to actively modulate how representations are formed, how information is routed, and how future trajectories are generated.

\subsubsection{UG-WIAE generator: probabilistic encoding with an uncertainty gate}

Given an input window $\mathbf{X}\in\mathbb{R}^{B\times F\times T}$, we retain the original convolutional encoding backbone to extract temporal features
\begin{equation}
\mathbf{h} = \phi(\mathbf{X}),
\end{equation}
and explicitly produce two branches:
\begin{equation}
\boldsymbol{\mu} = \psi_\mu(\mathbf{h}), 
\qquad 
\mathbf{u} = \psi_u(\mathbf{h}),
\end{equation}
where $\boldsymbol{\mu}$ represents a deterministic latent signal and $\mathbf{u}$ serves as a sample- and time-dependent proxy for uncertainty.

We further introduce a gating network that outputs an element-wise gate
\begin{equation}
\mathbf{g} = \sigma\!\left(\psi_g(\mathbf{h})\right), 
\qquad \mathbf{g}\in[0,1],
\end{equation}
which controls the strength of stochastic injection, i.e., how conservative or exploratory the model should be at each time step. A probabilistic latent (pseudo-innovation) is then constructed via a gated reparameterization:
\begin{equation}
\mathbf{z} = \boldsymbol{\mu} + \mathbf{g}\odot s(\mathbf{u})\odot \boldsymbol{\epsilon},
\qquad 
\boldsymbol{\epsilon}\sim p(\boldsymbol{\epsilon}),
\end{equation}
where $s(\cdot)$ is a positive scaling function that maps the uncertainty proxy to a stable noise scale, and $p(\boldsymbol{\epsilon})$ denotes a base noise distribution. As in WIAE, the desired prior structure of the latent variable (e.g., IID uniform) is enforced by the innovation discriminator during adversarial training. The role of the UG mechanism is to \emph{learn} where and how much noise should be injected, replacing fixed or manually tuned stochastic perturbations with uncertainty-adaptive control.

\paragraph{Uncertainty-aware attention: confidence-modulated routing}

In standard attention mechanisms, information aggregation is driven solely by pairwise similarity between queries and keys, which can lead to the uncritical amplification of spurious correlations under noise, distribution shift, or structural breaks. To address this limitation, UG-WIAE augments self-attention with an explicit \emph{uncertainty-aware confidence modulation}, allowing the model to attenuate unreliable interactions in a principled manner.

Given queries $\mathbf{Q}$, keys $\mathbf{K}$, and values $\mathbf{V}$, the conventional scaled dot-product attention is
\begin{equation}
\mathrm{Attn}(\mathbf{Q},\mathbf{K},\mathbf{V})
= \mathrm{Softmax}\!\left(\frac{\mathbf{Q}\mathbf{K}^\top}{\sqrt{d}}\right)\mathbf{V},
\end{equation}
where $d$ denotes the key dimension. This formulation implicitly assumes that similarity alone is sufficient to determine the reliability of information flow.

We instead introduce an uncertainty-dependent confidence gate that modulates attention scores. Let
\begin{equation}
\mathbf{S} = \frac{\mathbf{Q}\mathbf{K}^\top}{\sqrt{d}}
\end{equation}
denote the similarity matrix, and let $\mathbf{a}\in[0,1]$ be a confidence tensor inferred from uncertainty estimates associated with the representations (e.g., derived from latent variance or uncertainty proxies). The gated attention scores are defined as
\begin{equation}
\tilde{\mathbf{S}} = \mathbf{S} \odot \mathbf{a},
\end{equation}
leading to the uncertainty-aware attention output
\begin{equation}
\boldsymbol{\alpha} = \mathrm{Softmax}(\tilde{\mathbf{S}}),
\qquad
\mathrm{Output} = \boldsymbol{\alpha}\mathbf{V}.
\end{equation}

This formulation can be interpreted as a confidence-weighted routing mechanism: when uncertainty is high, the corresponding gates reduce the effective magnitude of attention scores, yielding smoother and more conservative aggregation; when uncertainty is low, the mechanism recovers standard similarity-driven attention. Importantly, this modulation does not alter the attention architecture itself and can be applied as a plug-in component to existing attention-based backbones. By explicitly incorporating uncertainty into the attention computation, UG-WIAE prevents low-confidence evidence from being disproportionately amplified, thereby improving robustness and interpretability in regimes characterized by noise, non-stationarity, or distributional shift.

\paragraph{Probabilistic decoding with gated distributional control}

At the decoding stage, we extend deterministic reconstruction to conditional distributional prediction. Given an intermediate representation $\mathbf{r}$, the decoder parameterizes a predictive distribution by jointly estimating its location and scale:
\begin{equation}
\boldsymbol{\mu}_Y = \rho_\mu(\mathbf{r}),
\qquad 
\log \boldsymbol{\sigma}_Y^2 = \rho_\sigma(\mathbf{r}),
\end{equation}
where $\rho_\mu$ and $\rho_\sigma$ are learnable decoding functions. This defines a conditional predictive family
\begin{equation}
p_\theta(\mathbf{Y}\mid \mathbf{r}) 
= \mathcal{N}\!\left(\boldsymbol{\mu}_Y,\;\mathrm{diag}\!\big(\boldsymbol{\sigma}_Y^2\big)\right).
\end{equation}

To couple predictive uncertainty with controllable conservatism, we introduce an output-side uncertainty gate $\mathbf{g}'\in[0,1]$, which modulates stochasticity at the sampling level. Specifically, predictive samples are generated via a gated reparameterization:
\begin{equation}
\hat{\mathbf{Y}} 
= \boldsymbol{\mu}_Y 
+ \mathbf{g}' \odot \boldsymbol{\sigma}_Y \odot \boldsymbol{\epsilon}',
\qquad 
\boldsymbol{\epsilon}' \sim p(\boldsymbol{\epsilon}'),
\end{equation}
where $p(\boldsymbol{\epsilon}')$ denotes a fixed base noise distribution (e.g., standard Gaussian).

This formulation preserves a well-defined probabilistic forecasting interface while allowing the effective dispersion of the predictive distribution to vary with uncertainty. When $\mathbf{g}'$ is small, sampling concentrates around the mean, yielding sharper and more conservative predictions; when $\mathbf{g}'$ increases, the model expresses greater variability, enabling robust scenario generation. Importantly, the gate operates on the distributional scale rather than altering the predictive mean, ensuring that uncertainty primarily governs dispersion instead of biasing central tendency.

\subsection{Experimental Setup}
All experiments were conducted on a single NVIDIA RTX 3080 (8GB VRAM). To ensure a fair efficiency comparison, all models were trained and evaluated under the same data split, input/output horizons, batch size, and early-stopping criterion. We report standard forecasting metrics as well as wall-clock time. Our UG-based model is consistently more efficient than alternatives, delivering an end-to-end runtime reduction of about 52\% compared to PatchTFT under the same evaluation protocol.

\end{document}